\AtBeginDocument{%
  }

\documentclass[dvipsnames,format=sigconf,anonymous=false,review=false]{acmart}

\copyrightyear{2026}
\acmYear{2026}
\setcopyright{cc}
\setcctype{by}
\acmConference[GECCO Companion '26]{Genetic and Evolutionary Computation Conference}{July 13--17, 2026}{San Jose, Costa Rica}
\acmBooktitle{Genetic and Evolutionary Computation Conference (GECCO Companion '26), July 13--17, 2026, San Jose, Costa Rica}
\acmDOI{10.1145/3795101.3814704}
\acmISBN{979-8-4007-2488-6/2026/07}

\usepackage{amssymb}
\usepackage{tabularx}
\usepackage{subcaption} 
\usepackage{caption}
\usepackage{hyperref}

\begin{document}

\title{Evolutionary Algorithm for Reservoir Learning and Yielding}



\author{Julien Testu}
\orcid{0009-0008-7833-2191}

\affiliation{%
  \institution{Inria}
  \institution{LaBRI, CNRS UMR 5800}
  \institution{IMN, University of Bordeaux, \\CNRS UMR 5293}
  \country{France}
}
\email{julien.testu@inria.fr}

\author{Pierrick Legrand}
\authornotemark[1]
\affiliation{%
  \institution{Bordeaux INP, ENSC}
  \institution{Inria}
  \institution{IMS, CNRS UMR 5218}
  \country{France}}
\email{pierrick.legrand@ensc.fr}

\author{Xavier Hinaut}
\authornote{Authors co-supervised this study.}
\affiliation{%
   \institution{Inria}
 \institution{IMN, University of Bordeaux, \\CNRS UMR 5293}
 \institution{LaBRI, CNRS UMR 5800}
 \country{France}
}
\email{xavier.hinaut@inria.fr}

\renewcommand{\shortauthors}{Testu et al.}

\begin{abstract}

Reservoir computing, a type of recurrent neural network, is a promising approach for temporal learning as it separates dynamic processing from the trained readout layer. However, classical Echo State Networks (ESNs) often require task-specific tuning of their architecture and hyperparameters to achieve good performance. This paper introduces EARLY (Evolutionary Algorithm for Reservoir Learning and Yielding), a framework designed to evolve both the topology and hyperparameters of multi-reservoir ESNs. Inspired by the modular organisation of the brain, EARLY encodes architectures as graph-based genomes and applies crossover, mutation, and selection to discover effective configurations. Our goal is to create both \emph{generic architectures}  and \emph{tasks inducing generalization}. The method is evaluated on temporal learning tasks from the CogScale dataset. Results show that evolved architectures outperform those obtained with random search on several tasks and exhibit structural differences depending on task difficulty: simpler tasks yield lightweight architectures, while more complex tasks favour richer modular organisations. These findings suggest that evolutionary search can help identify reusable reservoir structures for a broader range of temporal problems. The evolved architectures are further evaluated on a cross-situational learning dataset to assess their ability to adapt to new environments.
\end{abstract}

\ccsdesc[500]{Computing methodologies~Neural networks}
\ccsdesc[500]{Computing methodologies~Genetic algorithms}
\ccsdesc[300]{Computing methodologies~Supervised learning by regression}
\ccsdesc[300]{Computing methodologies~Supervised learning by classification}

\keywords{Evolutionary Algorithm, Reservoir Computing, Brain Connectivity, Multi-Task Training, Modularity, Generalisation, Time Series}

\maketitle

\section{Introduction}

The human brain can solve a wide variety of problems while consuming very little energy (approximately 20 W, comparable to two energy-saving light bulbs). Its modular organisation is believed to be the reason why it can solve and adapt to a wide variety of tasks~\cite{clune2013evolutionary}. Cortical columns are often described as generic brain structures rather than task-specific structures: their functional role is thought to depend on both their local microcircuitry and their pattern of connectivity.
This ability to reuse similar structures across multiple contexts is a remarkable example of generic architecture in nature. In machine learning such adaptability remains difficult to obtain: most models are designed for specific tasks and require substantial training before performing well in a new context. 

Reservoir computing offers an interesting direction because it separates the dynamic component, the reservoir, from a trainable readout layer, a single output layer that can be learned via linear regression. 
Thus, a reservoir transforms an input sequence into a rich internal state (used as a working memory \cite{strock2020robust}) with an output layer simple to train. Such structure alleviates the need for global weight optimization, reducing training cost, while also drawing inspiration from brain organization, where recurrent microcircuits such as cortical columns provide rich dynamics that are read out by downstream adaptive processes~\cite{seoane2019evolutionary}. In practice however, the performance of an Echo State Network on a particular task  depends on the choice of its architectures and hyperparameters \cite{jarvis2010extending}.
This work studies whether multi-reservoir architectures can improve this lack of adaptability and reduce the need for extensive hyperparameter tuning for each new task. More specifically, it addresses the following question: to what extent can evolving structures composed of multiple reservoirs promote modularity and improve generalisation across tasks?
In other words, our goal is to create both \emph{generic architectures} and \emph{tasks inducing generalization}. To explore this idea, this paper introduces EARLY (Evolutionary Algorithm for Reservoir Learning and Yielding), a graph-based evolutionary framework that enables the joint optimisation of both the topology and the hyperparameters of multi-reservoir Echo State Networks. 
The paper provides an empirical evaluation on temporal learning tasks, showing that the evolved architectures can outperform a random search baseline across several cognitive-like tasks. Finally, the structures will be tested on an unseen task of cross situational learning \cite{variengien2020journey}.

\section{Related Work}
Reservoir computing has been widely studied as a simple and efficient way to process sequential data. In Echo State Networks, the internal recurrent weights are fixed after random initialisation and only the readout is trained. This makes training efficient but also makes performance highly dependent on the choice of the reservoir hyperparameters and internal structure \cite{seoane2019evolutionary}. Previous work has explored how to tune these hyperparameters more efficiently and how to improve reservoir structures for specific tasks. Random search and task-oriented tuning procedures have shown that ESN performance can vary strongly depending on spectral radius, leak rate, and scaling choices \cite{hinaut2021hype}. Other works has explored biologically inspired reservoir structures to extend internal stability \cite{jarvis2010extending}.
In parallel, neuroevolution methods such as NEAT and HyperNEAT have shown that evolutionary search can successfully optimise neural structures by modifying both connectivity and model parameters \cite{stanley2002evolving, stanley2009hypercube}. 
These approaches motivated the design of EARLY. However, whereas NEAT evolves neural networks in which each node corresponds to a single neuron, the present framework treats nodes as reservoirs, thereby motivating the adoption of a novel method. While previous work has explored evolutionary optimisation of ESNs \cite{jiang2008supervised, basterrech2022evolutionary} and separately, the design of structured or multi-reservoir architectures \cite{manneschi2021exploiting}, the joint evolution of both topology and hyperparameters in multi-reservoir ESNs remains largely unexplored.

\section{Method}

\subsection{Multi-Reservoir Echo State Networks}

A classical Echo State Network is composed of an input layer, a reservoir and a readout layer. The reservoir projects the input into a higher-dimensional dynamic space called the reservoir's state and the readout learns a mapping from those states to the target output. In this work, the reservoir is extended into a set of interacting reservoirs. Each reservoir can have different dynamic properties and different connections to other reservoirs and to the readout. This allows the model to combine multiple temporal transformations instead of relying on a single recurrent block.

\begin{figure}[h]
  \centering
  \includegraphics[width=\linewidth]{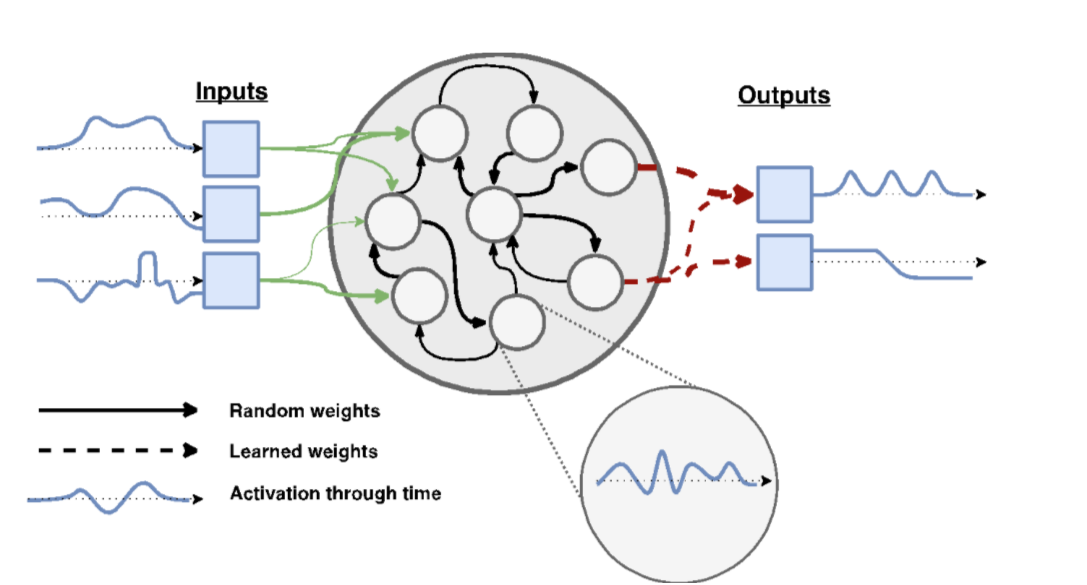}
  \caption{Single-reservoir Echo State Network. 
  Green and black connections are random and kept fixed, 
  while red connections are trained (i.e. the \emph{readout}). Image from \cite{juven2020cross}.}
  \label{fig:esn}
\end{figure}

\subsection{Genome Representation}

Each candidate architecture is represented as a matrix-based genome (see table~\ref{tab:genome}). The first part encodes the connectivity between input, reservoirs and readout as an adjacency matrix. The second part stores the hyperparameters of each reservoir. In this representation, a direct connection and a feedback connection are distinguished which allows the model to represent richer temporal interactions. For each reservoir the genome stores the amount of units, leak rate, input scaling, spectral radius, feedback scaling, input connectivity, reservoir connectivity and feedback connectivity. This representation makes it possible to evolve both the global architecture and the local dynamics of each module in a unified way.

\subsection{Evolutionary Process}
EARLY follows a steady-state evolutionary process in which an initial population of candidate architectures is randomly generated and evaluated. At each generation, two parents are selected through repeated two-player tournament selection, introducing selection pressure while preserving diversity. Offspring are produced by crossover and mutation. Crossover mixes adjacency-matrix columns according to a random ratio which vertically cuts each parents. The offspring will inherit the left side of parent 1's genome and the right side of parent 2's genome (see figure~\ref{fig:crossover}). Hyperparameters are recombined independently for each row using three crossover points, yielding alternating segments from both parents. Mutation can add, remove, or modify direct and feedback connections, insert or delete reservoirs, and perturb hyperparameters using integer changes for discrete variables and bounded perturbations for continuous ones.
After evaluation, offspring are merged with the parent population, ranked by fitness, and filtered through elitist truncation to retain only the best individuals at a fixed population size.

\begin{table}
\small

\begin{tabularx}{\columnwidth}{lcccccc}
\toprule
 & INPUT & RES1 & RES2 & RES3 & RES4 & READOUT  \\
\midrule
INPUT    & $\varnothing$ & 1 & 0 & 1 & 1 & 1  \\
RES1     & $\varnothing$ & $\varnothing$ & 0 & 1 & 0 & 1  \\
RES2     & $\varnothing$ & 2 & $\varnothing$ & 0 & 0 & 0  \\
RES3     & $\varnothing$ & 2 & 1 & $\varnothing$ & 1 & 1  \\
RES4     & $\varnothing$ & 2 & 0 & 0 & $\varnothing$ & 1  \\
READOUT  & $\varnothing$ & 2 & 0 & 0 & 2 & $\varnothing$ \\
\bottomrule
\end{tabularx}

\vspace{4pt}

\begin{tabularx}{\columnwidth}{lcccccccc}
\toprule
 & U & LR & IN\_S & SR & FB\_S & IC & RC & FB \\
\midrule
INPUT    & 0   & 0    & 0   & 0   & 0   & 0   & 0   & 0 \\
RES1     & 39 & 0.3  & 3.5 & 2.8 & 0.01 & 0.1 & 0.3 & 0.7 \\
RES2     & 40 & 0.6  & 0.4 & 0.03 & 0.4 & 0.3 & 0.1 & 0.3 \\
RES3     & 86 & 0.01 & 2.4 & 1.3 & 0.9 & 0.8 & 0.2 & 0.1 \\
RES4     & 65 & 0.9  & 1.8 & 3.2 & 1.0 & 0.5 & 0.9 & 0.009 \\
READOUT  & 0   & 0    & 0   & 0   & 0   & 0   & 0   & 0 \\
\bottomrule
\end{tabularx}
\caption{Genome of a multi-reservoir architecture}
\label{tab:genome}
\end{table}

\subsection{Training and Fitness}
For each individual, the corresponding multi-reservoir ESN is instantiated and only the readout layer is trained. This training step uses ridge regression, with the regularization parameter selected by cross-validation. An individual’s fitness is defined by its task performance, with lower error indicating better fitness. 
To reduce optimisation time, architectures were evaluated using a single task seed. This seed was selected as the one yielding median performance across 50 task seeds for a randomly initialised individual, ensuring it was representative of the task difficulty. Since reservoir weights remain randomly initialised, the final fitness was computed as the mean performance over five reservoir initialisations using that fixed task seed.

\subsection{Tasks}
The evaluation focuses on \href{https://github.com/Naowak/CogScale}{\emph{CogScale}} \cite{bendiouis2026benchmark}, a benchmark of sequential tasks designed to assess memory and temporal reasoning across various settings, including forecasting, postcasting, pattern completion, copy tasks, bracket matching, sorting, and adding problems. This diversity makes it a suitable testbed for evaluating both predictive performance and the structural properties of the evolved reservoirs. We also evaluate generalisation in a new, unseen Cross-Situational Learning (CSL) task  \cite{variengien2020journey}. In CSL, the model must infer word–meaning associations from repeated co-occurrences between input sentences and ambiguous visual scenes, rather than from a fully explicit target. More specifically, the model receives a sentence word by word and must reconstruct a scene representation even though the target scene may contain information not directly stated in the observation. This makes CSL a useful test of generalisation beyond standard supervised temporal benchmarks, the vocabulary size of the task is set at 20.

\subsection{Experiments}
The main comparison is between EARLY and a random search baseline. For each CogScale task, both methods generate a task-specific architecture. The objective is to determine whether EARLY can find lower-error solutions than random search under an identical budget of 50,100 evaluated architectures, while also analysing how the resulting structures vary across tasks. EARLY evaluates 100 initial random individuals followed by 50 offspring per generation over 1,000 generations, whereas random search evaluates 50,100 independent random architectures. To assess generalisation, the resulting architectures are evaluated beyond their original optimisation task. Each architecture is first tested on all other CogScale tasks to measure how well task-specific structures generalise across the benchmark. Architectures obtained with both EARLY and random search are then also evaluated on the CSL task, allowing a direct comparison of how well evolved and randomly sampled structures adapt to a new unseen environment.

\section{Results}

\subsection{Architecture search comparison}

Table \ref{tab:results_STREAM} shows that EARLY consistently outperforms random search on all CogScale tasks under the same evaluation budget. The improvement is visible on both simpler forecasting tasks and more demanding memory-based tasks, with particularly large gains on the adding problem, bracket matching, simple copy, and discrete pattern completion. On average, the mean error decreases from 0.214 with random search to 0.129 with EARLY, indicating that evolutionary optimisation is substantially more effective than pure random sampling for identifying high-performing architectures. 

\begin{table}[h!]
\centering
\begin{tabular}{lcc}
\hline
\textbf{Task Name} & \textbf{Random Search} &  \textbf{EARLY} \\
\hline
Chaotic Forecasting & 5.7e-05 & \textbf{6.305e-06} \\
Sinus Forecasting & 0.0005 & \textbf{5.82e-05} \\
Continuous Postcasting & 0.0048 & \textbf{0.0022} \\
Discr. Pattern Completion & 0.107 & \textbf{0.058} \\
Cont. Pattern Completion & 0.038 & \textbf{0.021} \\
Bracket Matching & 0.149 & \textbf{0.065} \\
Simple Copy & 0.474 & \textbf{0.213} \\
Selective Copy & 0.464 & \textbf{0.42} \\
Adding Problem & 0.388 & \textbf{0.07} \\
Sorting Problem & 0.5094 & \textbf{0.44} \\
\hline
\textbf{Mean} & 0.214 & \textbf{0.129} \\
\hline
\end{tabular}
\caption{Errors across tasks. (Lower is better)}
\label{tab:results_STREAM}
\end{table}

These results suggest that jointly optimising connectivity and reservoir parameters helps discover more suitable multi-reservoir organisations for temporal processing. This is consistent with previous work showing that structured reservoir connectivity can improve stability and performance \cite{jarvis2010extending}.

\subsection{EARLY generalization across tasks}

\begin{figure}[h]
\centering
\includegraphics[width=1\linewidth]{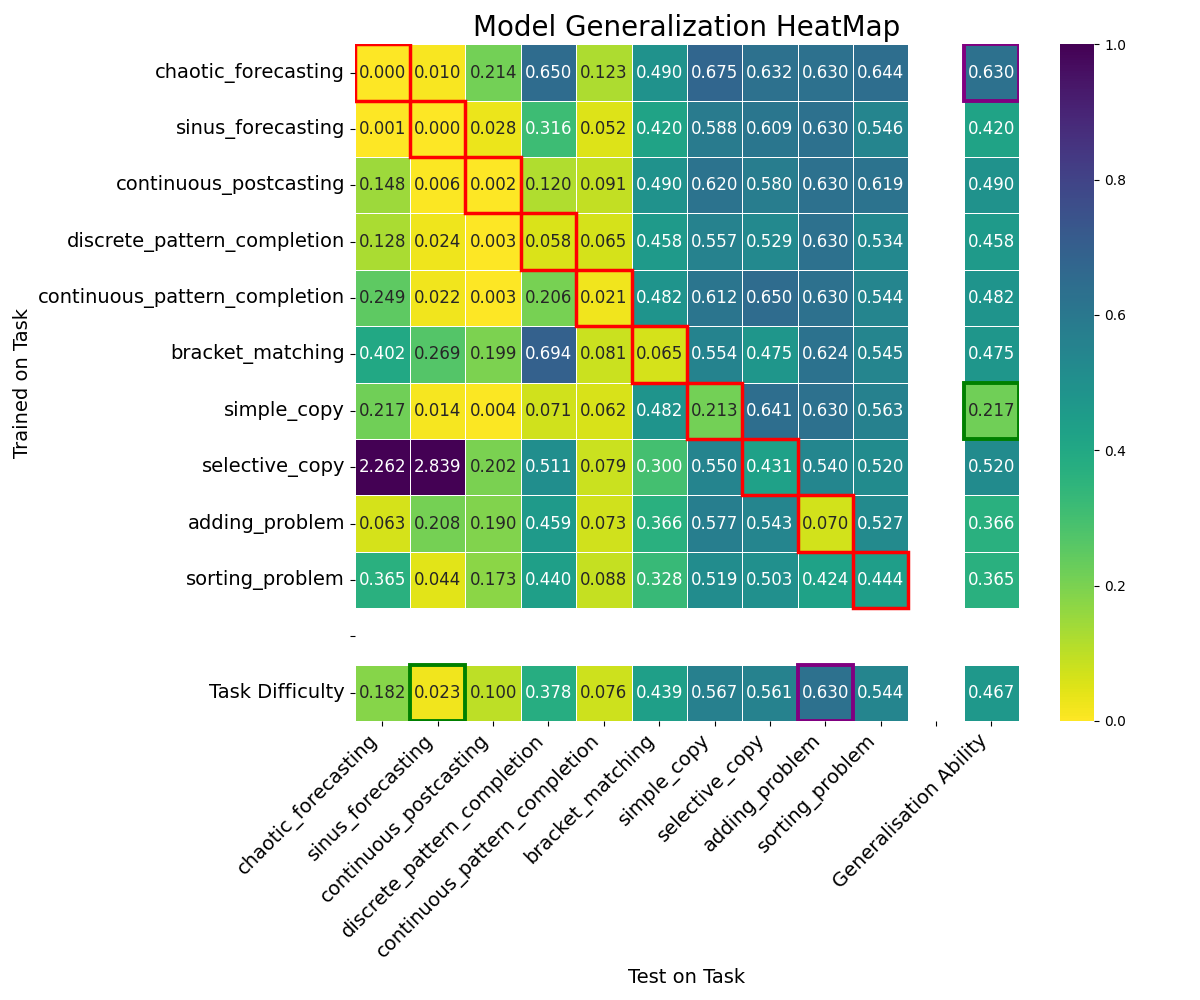}
\caption{Cross-task error heatmap for EARLY. (Lower is better.) Rows correspond to source tasks and columns to target tasks.  Medians of columns (\emph{Task Difficulty}) and rows (\emph{Generalization Ability}) are separated from the matrix.}
\label{fig:hm_compare}
\end{figure}

Figure \ref{fig:hm_compare} shows the cross-task error heatmap for all tasks of CogScale dataset. 
We can see that some tasks induce evolved architectures that perform better on several tasks. \emph{Simple copy} task is clearly inducing the best generalization among all tasks; \emph{adding problem} and \emph{sorting problem} also obtain good results. Thus, we can say that these tasks (through EARLY) have good \emph{transfer capabilities} by themselves, which already induce fair \emph{generic architectures}.
Moreover, in the upper left corner of the heatmap, we see a low-error block: this suggests a sub-group of tasks sharing similar requirements.
By comparison, the Random Search (RS) heatmap, provided in the Appendix, shows weaker and noisier \emph{transfer capabilities}, with several source tasks yielding substantially larger errors outside their original regime. In particular, \emph{simple copy} is not among the best-performing tasks under RS. These results indicate that the observed \emph{transfer capability} arises not from the tasks alone, but from their combination with EARLY. The fact that evolutionary search yields both stronger source-task performance and more coherent transfer behaviour is consistent with previous work showing that optimizing reservoir structure can uncover more useful dynamical regimes than unguided sampling \cite{jiang2008supervised,basterrech2022evolutionary}. More generally, it supports the view that evolution can favour computational organizations with utility beyond the conditions in which they were originally selected \cite{seoane2019evolutionary}. The stronger reuse observed for some EARLY architectures further suggests that modular organization may facilitate adaptation across tasks and environments \cite{clune2013evolutionary}.

\subsection{Architecture transferability on CSL}

Table \ref{tab:csl} shows that architectures evolved with EARLY adapt better on average to the unseen CSL setting than architectures obtained by random search. EARLY achieves a lower mean validity error, decreasing the average validity error from 0.3194 to 0.1941. The advantage is visible for most source tasks, especially for continuous postcasting, pattern completion, bracket matching, and sorting. 
Random search remains better only for a small number of cases, notably chaotic forecasting and adding problem, while sinus forecasting yields nearly identical results for both methods. An explanation for chaotic and sinus forecasting is that those tasks are relatively simple, like the resulting evolved models, and thus produce less generalisable architectures.
These results suggest that the structural patterns introduced by evolutionary optimisation are not only beneficial on the original CogScale tasks but can also produce architectures that transfer more effectively to a different and more language-grounded setting.
\begin{table}[h]
\centering

\begin{tabular}{lcc}
\hline
\textbf{Task Individual} & \textbf{RS Valid} & \textbf{EARLY Valid} \\
\hline
Chaotic Forecasting & \textbf{0.679}  & 0.932 \\
Sinus Forecasting & \textbf{0.332}  & 0.334 \\
Continuous Postcasting & 0.4404  & \textbf{0.141}  \\
Discrete Pattern Completion & 0.1514  & \underline{\textbf{0.0084*}}  \\
Continuous Pattern Completion & 0.3822  & \textbf{0.128} \\
Bracket Matching & 0.9181 & \textbf{0.22}  \\
Simple Copy & 0.0820*  & \textbf{0.0495*}  \\
Selective Copy & 0.1101  & \textbf{0.0487*}  \\
Adding Problem & \textbf{0.0319*}  & 0.044*  \\
Sorting Problem & 0.0673*  & \textbf{0.035*}  \\
\hline
\textbf{Mean} & 0.3194  & \textbf{0.1941}  \\
\hline
\end{tabular}

\caption{Performance comparison across on the CSL dataset.
\underline{ } indicates best error.
* indicates errors $< 0.1$. }
\label{tab:csl}
\end{table}

These results support the view that evolutionary search can produce reservoir organisations that remain effective beyond the task on which they were optimised \cite{seoane2019evolutionary,jiang2008supervised,basterrech2022evolutionary}. They are also in line with previous findings suggesting that modular and structured systems are better suited to adaptation in novel environments \cite{clune2013evolutionary}.

\section{Discussion}

One of our long term goal is to obtain \emph{generic architectures} trained on \emph{tasks inducing generalization} by evolving multi-reservoir architectures. This study show interesting results going in this direction. 
First, we show that evolutionary optimisation with EARLY is more effective than random search for discovering high-performing multi-reservoir ESN architectures. 
On the CogScale dataset, EARLY consistently achieves lower error under the same evaluation budget, indicating that evolutionary search discovers more effective structural patterns than random search.
The cross-task analysis further shows that evolved architectures are not purely task-specific: 
simple tasks tend to produce architectures that generalize well on other simple tasks;
however, more difficult tasks induce architectures that are not as good on the simpler tasks, but they have a greater transfer capability on other tasks.
Comparing EARLY and RS heatmaps suggests over-specialisation in EARLY on selective copy and chaotic forecasting: although it achieves lower task-specific error, its generalisation is worse (resp. median 0.52 vs. 0.38, and 0.63 vs. 0.547). This indicates that these tasks favour specialisation over transfer, leading to high performance but limited generalisation. 
This trade-off between specialisation and reuse also appears when we evaluate the architectures on CSL data, for which none of the architectures have been evolved for. Architectures that were evolved on \emph{discrete pattern completion}, \emph{simple copy}, \emph{selective copy}, \emph{adding problem} and \emph{sorting problem} yield the CSL scores (below 0.1). 
As shown in the heatmap in Fig.~\ref{fig:hm_compare}, these tasks are also among the most difficult and require more complex architectures. 
This suggests that reusable generic architectures are more likely to emerge from sufficiently challenging tasks, as these encourage the development of structures able to support richer temporal dynamics and generalize beyond their original training environment. 
Overall, these results support the idea that reservoir organisation is central to both performance and adaptability, while indicating that multi-reservoir structures may serve as an effective foundation for transfer across tasks.

The full code is available at \href{https://github.com/Julien06T/Evolutionary-Algorithm-For-Reservoir-Learning-And-Yielding/blob/main/README.md}{\emph{project Repository}}. 
Future work will explore broader benchmarks and investigate whether promoting modularity by penalizing connectivity, improves generalisation \cite{clune2013evolutionary}.
In particular, we aim to identify the minimal set of \emph{tasks inducing generalization} that can support a wide range of problems, with the goal of obtaining \emph{generic architectures} at minimal computational cost.
Finally, we plan to perform meta-learning~\cite{leger2024evolving} using this minimal set of tasks.

\bibliographystyle{ACM-Reference-Format}
\bibliography{references}

\appendix
\section{Research Methods appendix}

\subsection{Supplementary info on genome validity correction}
When necessary, corrective connections are added so that every genome can be converted into a functional ESN. The procedure enforces valid input-to-reservoir and reservoir-to-readout connectivity, removes invalid self-connections, and ensures that cycles include enough feedback links to preserve a valid computation order.

\begin{figure}[h!]
\centering
\includegraphics[width=0.7\linewidth]{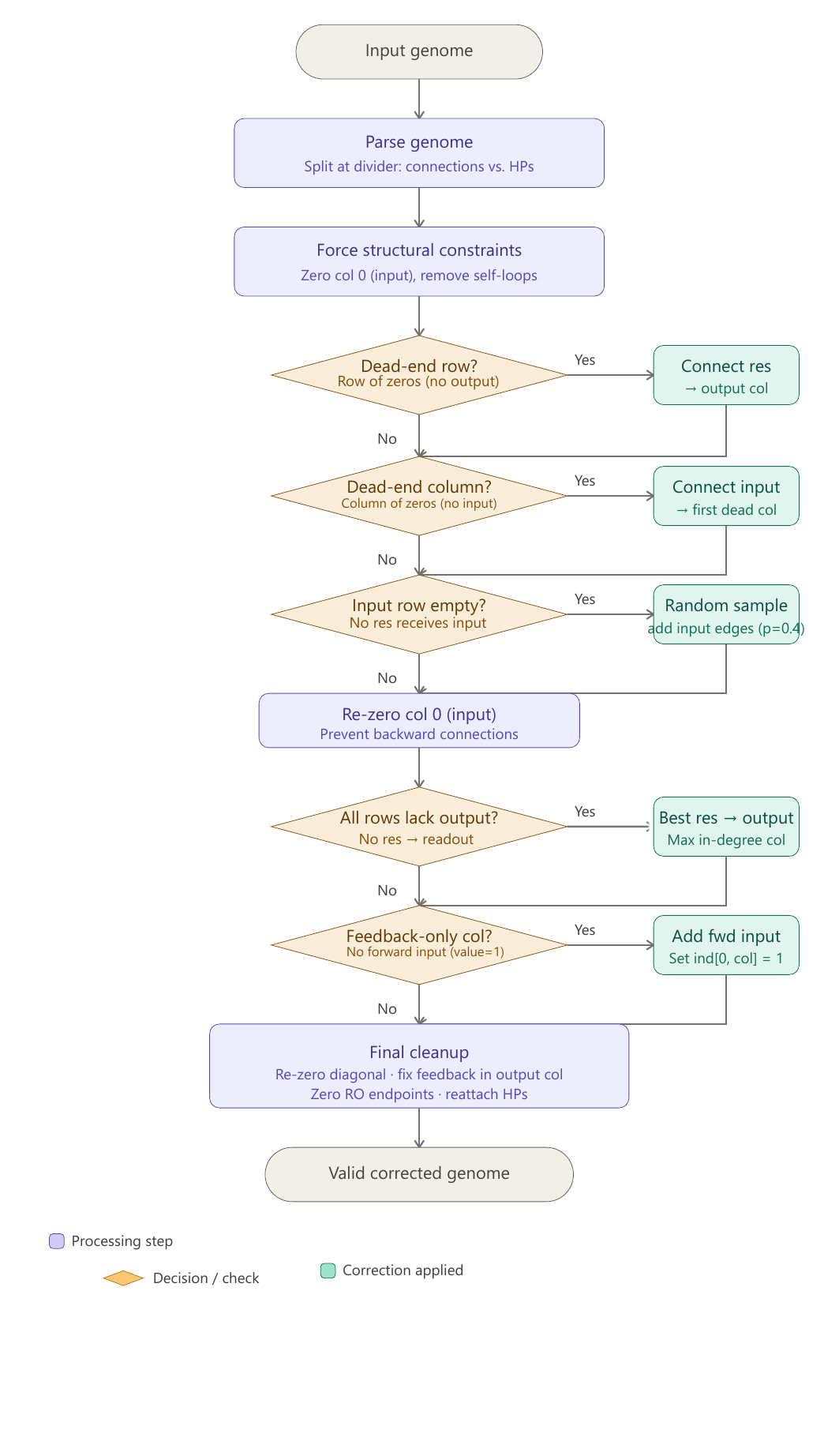}
\caption{Flowchart of the validity function}
\label{fig:flowchart_valid_func}
\end{figure}

\subsection{Crossover scheme}
\begin{figure}[h!]
\centering
\includegraphics[width=0.4\linewidth]{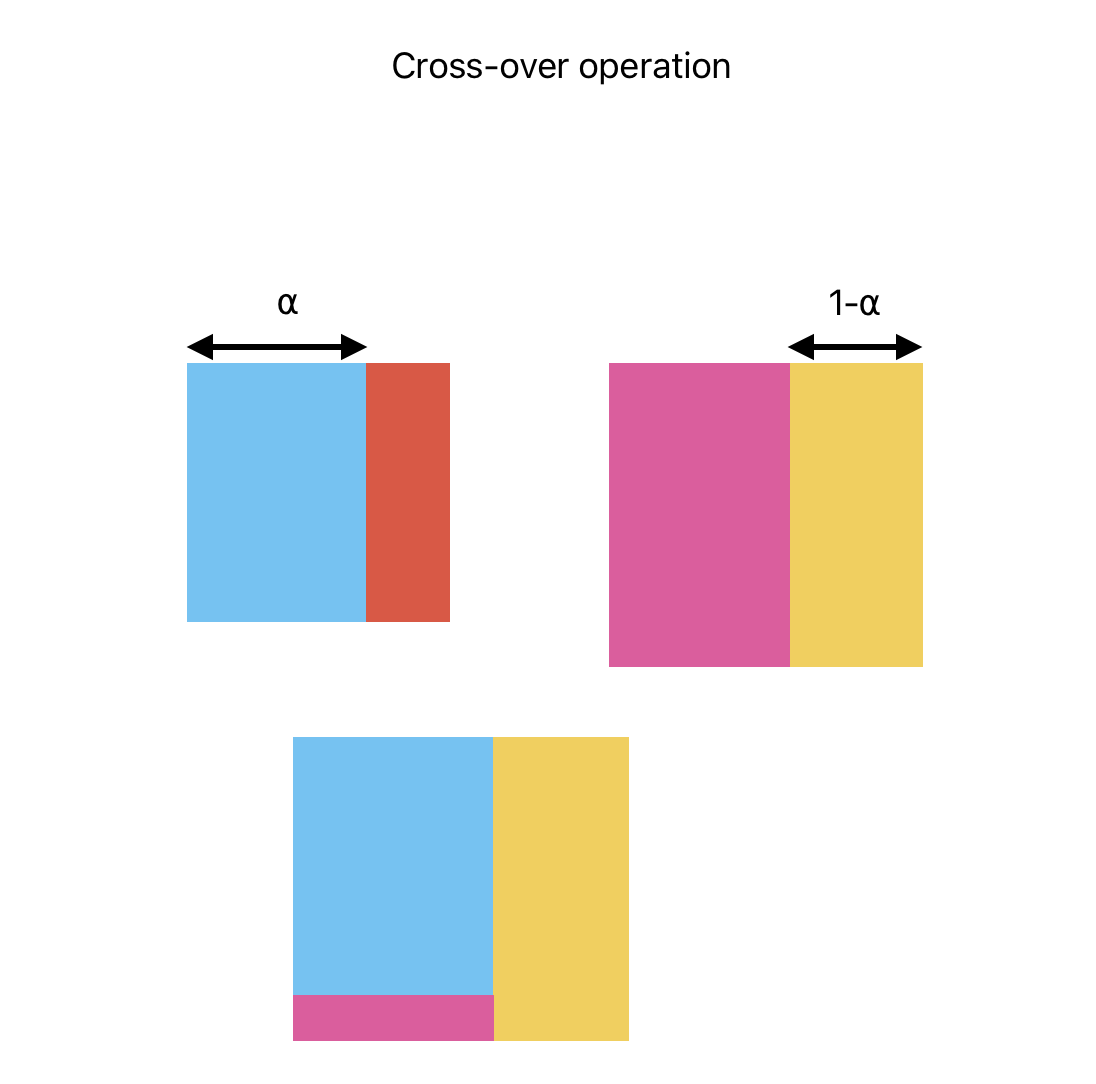}
\caption{Crossover scheme }
\label{fig:crossover}
\end{figure}
$\alpha$ represents the random ratio selected at each crossover steps. We take the first $\alpha \%$ of the first parent and the $1-\alpha \%$ of the second parent. If the parents are of different size, we complete the missing rows of the offspring with the rows of the largest parent in order to maintain the NxN dimension of the adjacency matrix, which is needed to ensure no dead-branch or start-nodes which are not of type input. 

\subsection{Cortical Columns Architecture}
\begin{figure}[h!]
\centering
\includegraphics[width=0.7\linewidth]{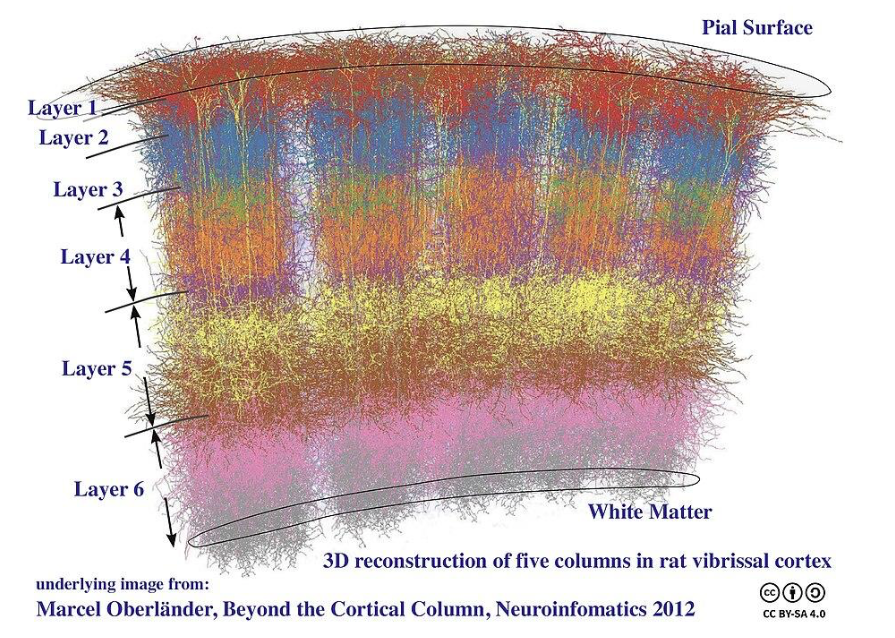}
\caption{Six-layer cortical column. Upper and middle layers (L2/3–L4) are linked to processing that can be reused across tasks, while deeper layers (L5–L6) are more task-specific and related to output and feedback. Image from Oberlaender et al.}
\label{fig:cortical_column}
\end{figure}

\textbf{Cortical columns} (see Figure~\ref{fig:cortical_column}\footnote{Image from Oberlaender, M., Narayanan, R., Egger, R., Meyer, H., Baltruschat, L., Dercksen, V., ... \& Sakmann, B. (2014). Beyond the Cortical Column-Structural Organization Principles in Rat Vibrissal Cortex. In Front. Neuroinform. Conference Abstract: 5th INCF Congress of Neuroinformatics.  (Vol. 52). }) are often described as generic brain structures rather than task-specific structures: their functional role is thought to depend on both their local microcircuitry and their pattern of connectivity to other areas.
A cortical column can be viewed as a local vertical microcircuit spanning the cortical layers, embedded in a larger hierarchy of cortical areas. In the classical view, feedforward inputs from lower-level areas mainly target layer 4 and arise predominantly from supragranular layers (layers 2/3), whereas feedback projections from higher-level areas arise mainly from infragranular layers (layers 5/6) and avoid layer 4. However, Markov and Kennedy\footnote{Markov, N. T., \& Kennedy, H. (2013). The importance of being hierarchical. Current opinion in neurobiology, 23(2), 187-194.} emphasize that this laminar organization is more complex, with distinct feedforward and feedback counter-streams present in both superficial and deep cortical compartments. 

\newpage

\subsection{EARLY vs Random Search : Optimisation Curve}

\begin{figure}[h]
\centering
\includegraphics[width=0.7\linewidth]{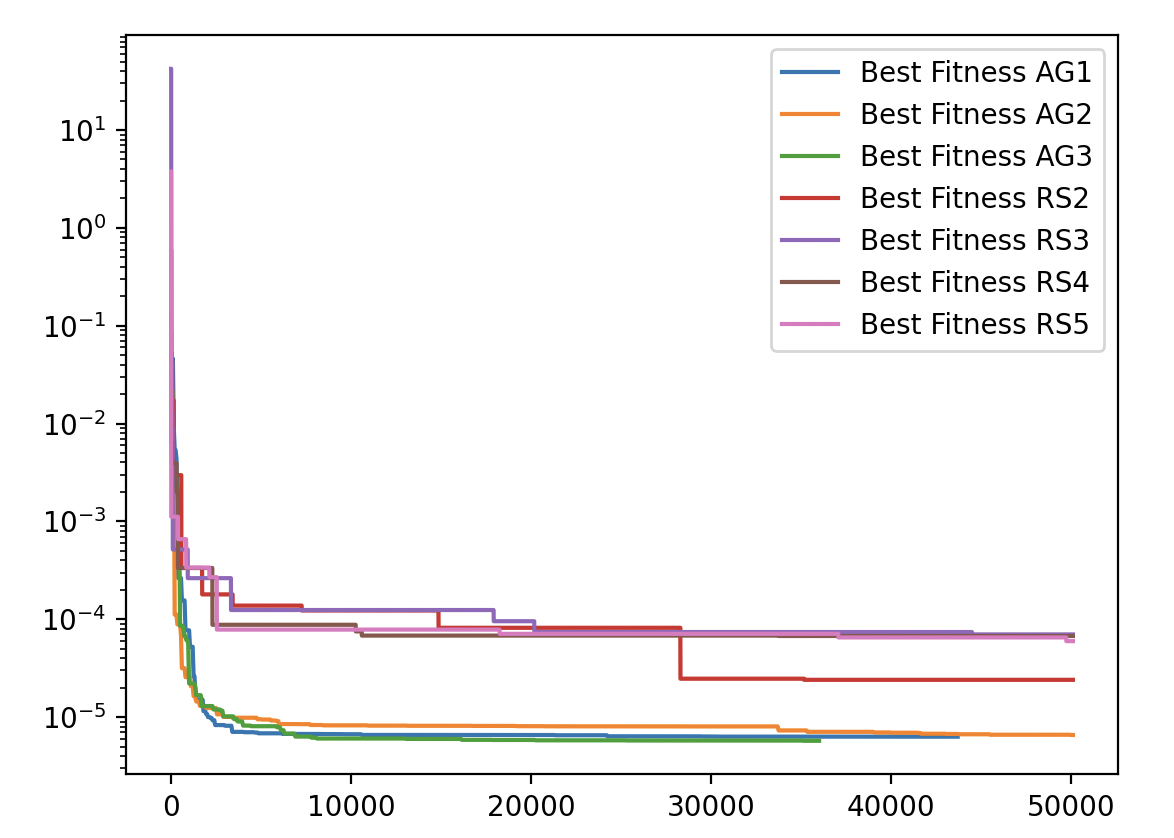}
\caption{Chaotic forecasting convergence. Due to time constraints, this is the only task where multiple EARLY runs where conducted.}
\label{fig:chaotic_convergence_curve}
\end{figure}

\begin{figure}[h]
\centering
\includegraphics[width=0.4\linewidth]{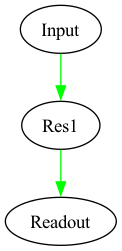}
\caption{Chaotic forecasting EARLY individual}
\label{fig:chaotic_ind}
\end{figure}

\begin{figure}[h]
\centering
\includegraphics[width=0.7\linewidth]{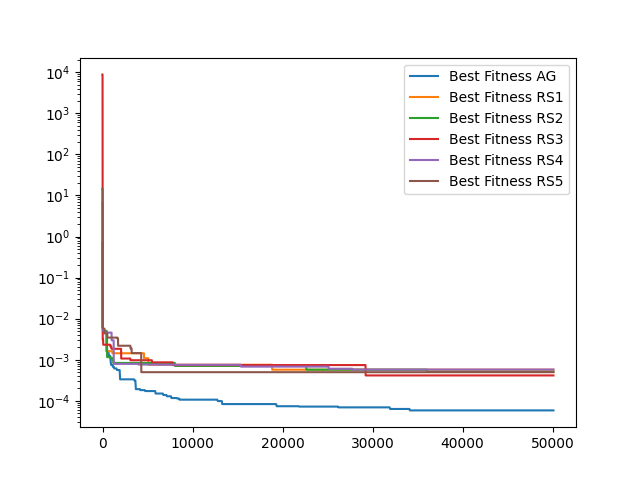}
\caption{Sinus forecasting convergence curve. 50100 different models and evaluations.}
\label{fig:sinus_convergence_curve}
\end{figure}

\begin{figure}[h]
\centering
\includegraphics[width=0.4\linewidth]{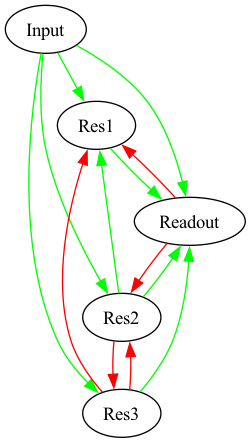}
\caption{Sinus forecasting EARLY individual}
\label{fig:sinus_ind}
\end{figure}

\begin{figure}[h]
\centering
\includegraphics[width=0.7\linewidth]{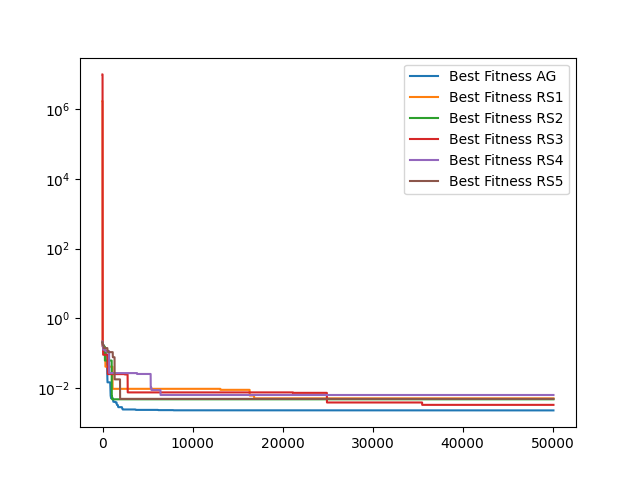}
\caption{Continuous postcasting convergence curve}
\label{fig:cont_post_conv}
\end{figure}

\begin{figure}[h]
\centering
\includegraphics[width=0.4\linewidth]{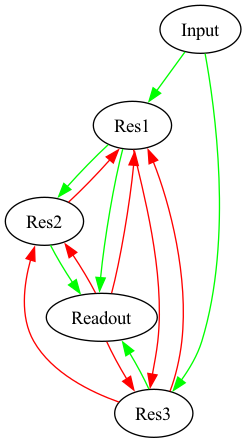}
\caption{Continuous postcasting EARLY individual}
\label{fig:cont_post_ind}
\end{figure}

\begin{figure}[h]
\centering
\includegraphics[width=0.7\linewidth]{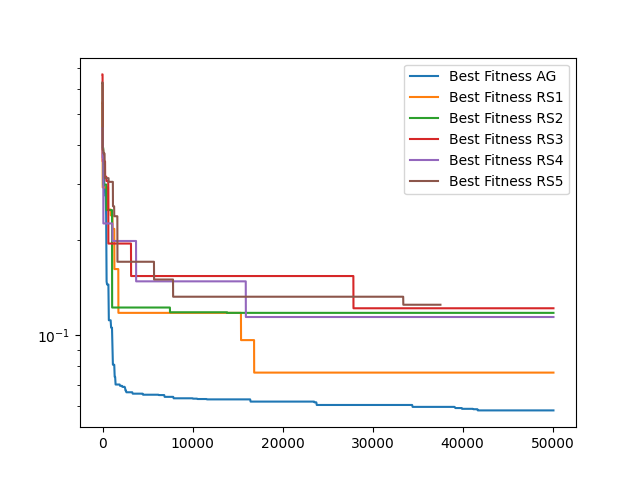}
\caption{Discrete pattern completion convergence curve}
\label{fig:disct_pat_curve}
\end{figure}

\begin{figure}[h]
\centering
\includegraphics[width=0.4\linewidth]{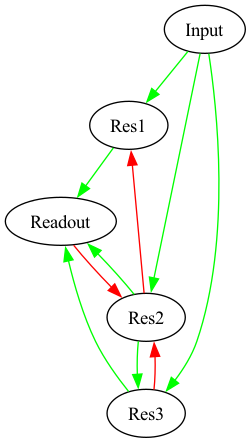}
\caption{Discrete Pattern Completion EARLY individual}
\label{fig:disct_pat}
\end{figure}

\begin{figure}[h]
\centering
\includegraphics[width=0.7\linewidth]{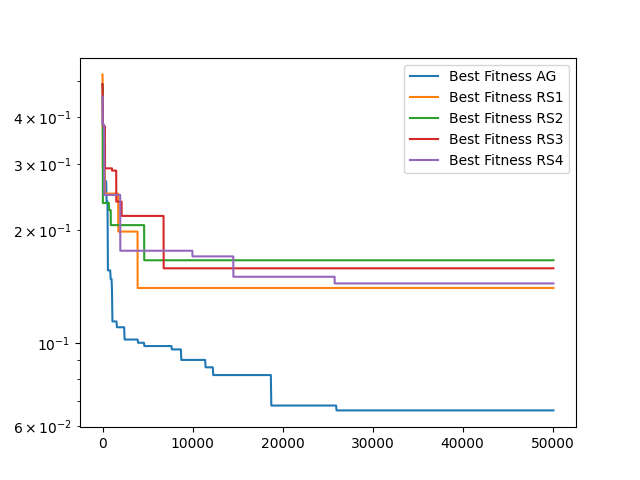}
\caption{Bracket matching convergence curve}
\label{fig:bm_curve}
\end{figure}

\begin{figure}[h]
\centering
\includegraphics[width=0.4\linewidth]{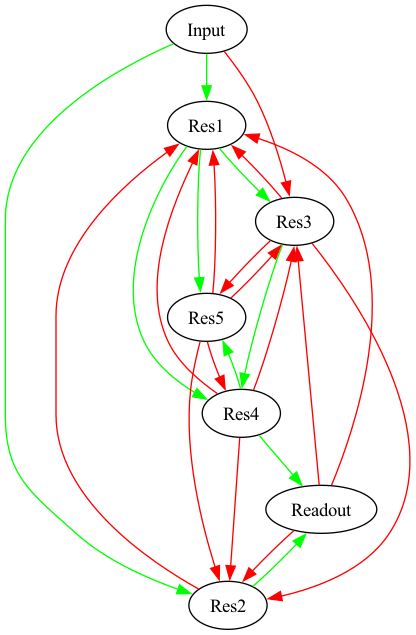}
\caption{Bracket Matching EARLY individual}
\label{fig:bm_ind}
\end{figure}

\begin{figure}[h]
\centering
\includegraphics[width=0.7\linewidth]{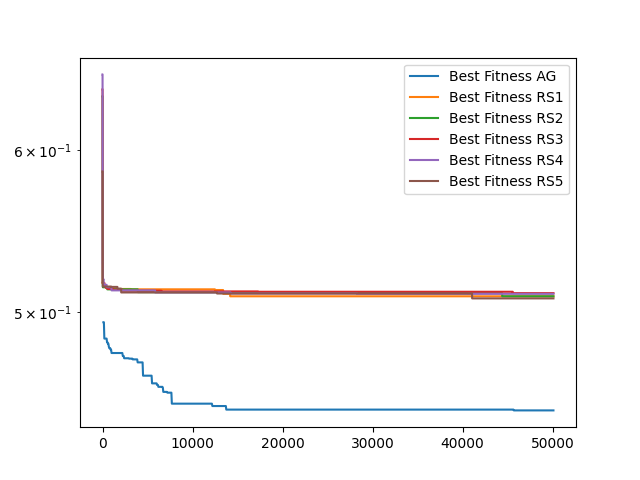}
\caption{Sorting problem convergence curve}
\label{fig:sp_curve}
\end{figure}

\begin{figure}[h]
\centering
\includegraphics[width=0.4\linewidth]{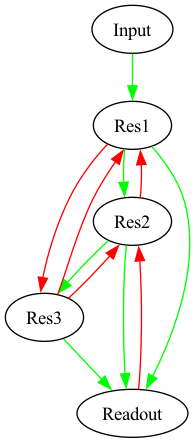}
\caption{Sorting problem EARLY individual}
\label{fig:sp_ind}
\end{figure}

\begin{figure}[h]
\centering
\includegraphics[width=0.7\linewidth]{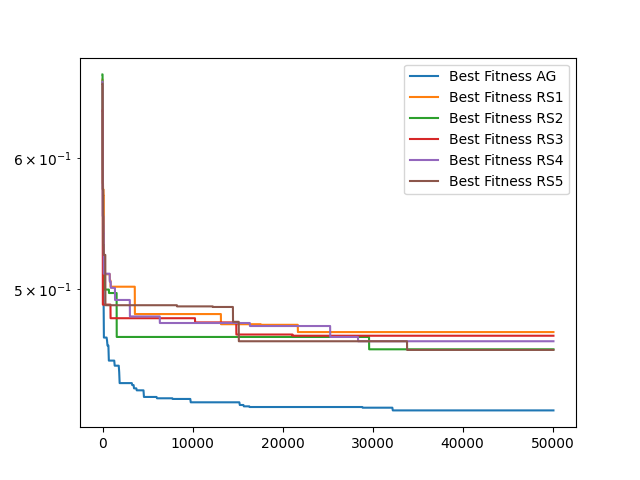}
\caption{Selective copy convergence curve. 50100 different models and evaluations.}
\label{fig:selective_copy_convergence_curve}
\end{figure}

\begin{figure}[h]
\centering
\includegraphics[width=0.4\linewidth]{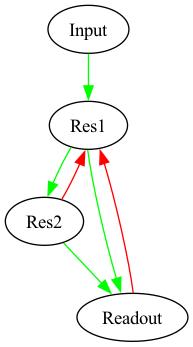}
\caption{Selective copy EARLY individual}
\label{fig:selective_copy_individual}
\end{figure}

\begin{figure}[h]
\centering
\includegraphics[width=0.9\linewidth]{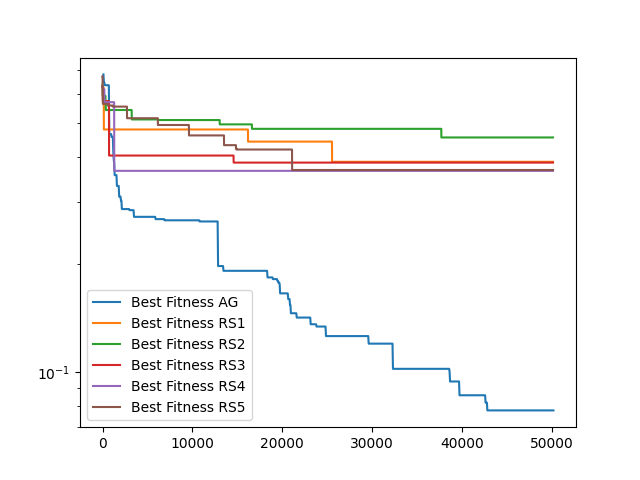}
\caption{Adding problem convergence curve.}
\label{fig:adding_problem_convergence_curve}
\end{figure}

\begin{figure}[h]
\centering
\includegraphics[width=0.6\linewidth]{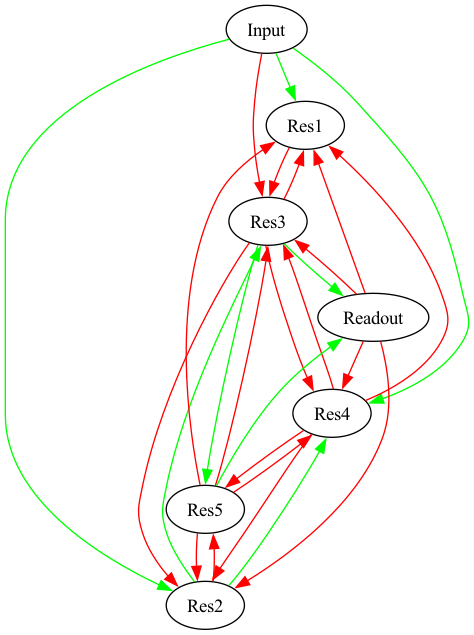}
\caption{Adding problem EARLY individual}
\label{fig:adding_problem_individual}
\end{figure}

\begin{figure}[h]
\centering
\includegraphics[width=0.9\linewidth]{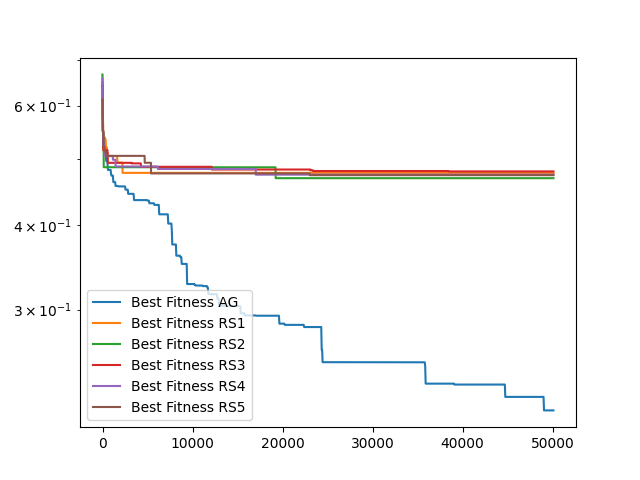}
\caption{Simple copy convergence curve.}
\label{fig:simple_copy_convergence_curve}
\end{figure}

\begin{figure}[h]
\centering
\includegraphics[width=0.6\linewidth]{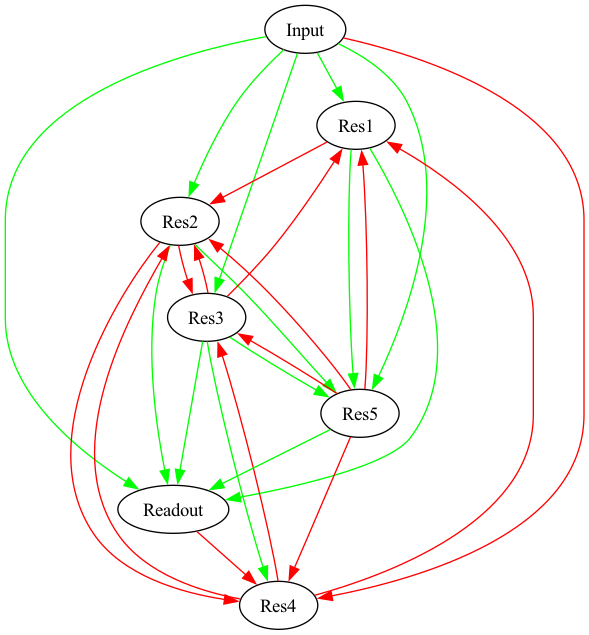}
\caption{Simple copy EARLY individual}
\label{fig:simple_copy_individual}
\end{figure}

\clearpage
\section{CogScale Tasks}

\begin{figure}[h]
\centering
\includegraphics[width=0.9\linewidth]{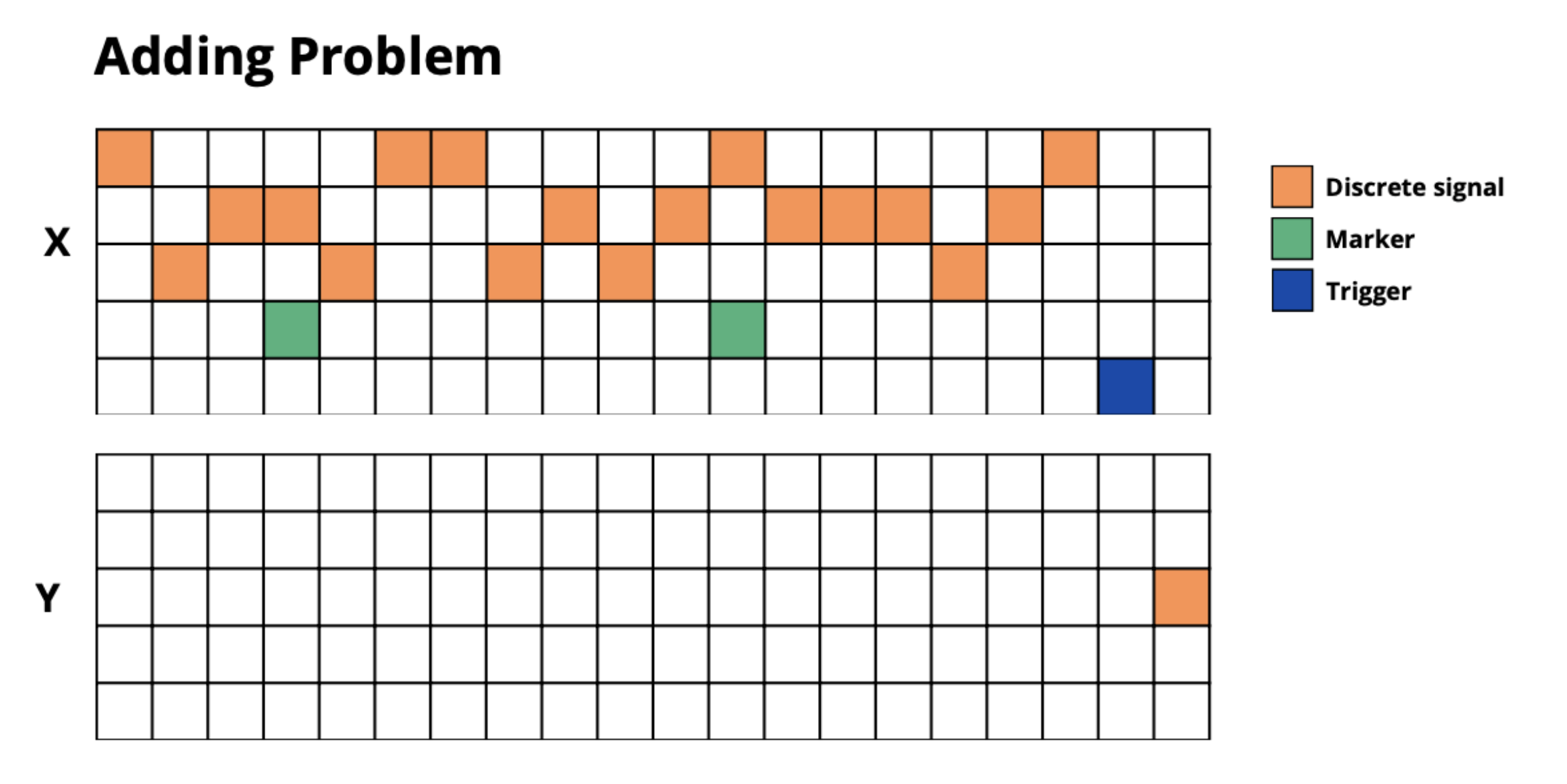}
\caption{Adding problem}
\label{fig:adding_problem}
\end{figure}

\begin{figure}[h]
\centering
\includegraphics[width=0.9\linewidth]{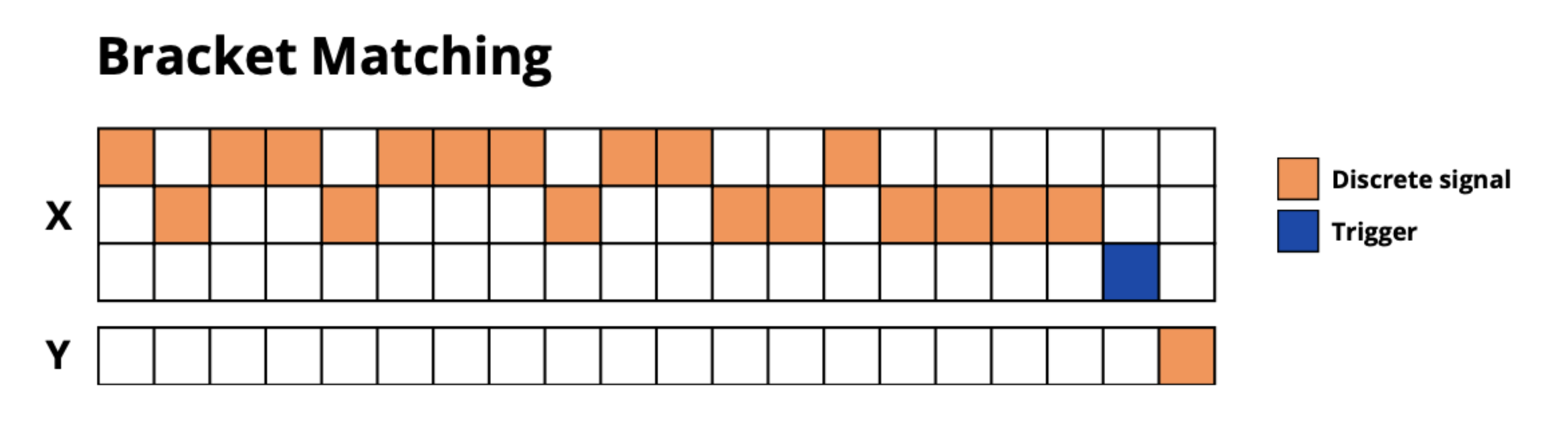}
\caption{Bracket matching}
\label{fig:bracket_matching}
\end{figure}

\begin{figure}[h]
\centering
\includegraphics[width=0.9\linewidth]{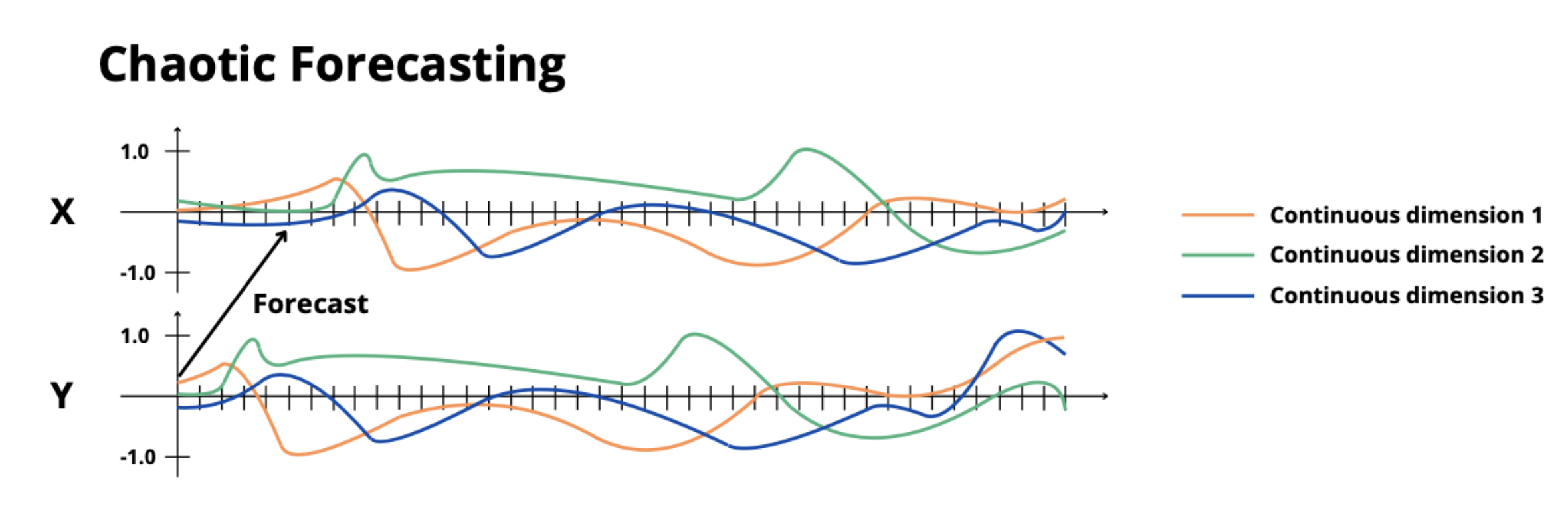}
\caption{Chaotic Forecasting}
\label{fig:chaotic_forecasting}
\end{figure}

\begin{figure}[h]
\centering
\includegraphics[width=0.9\linewidth]{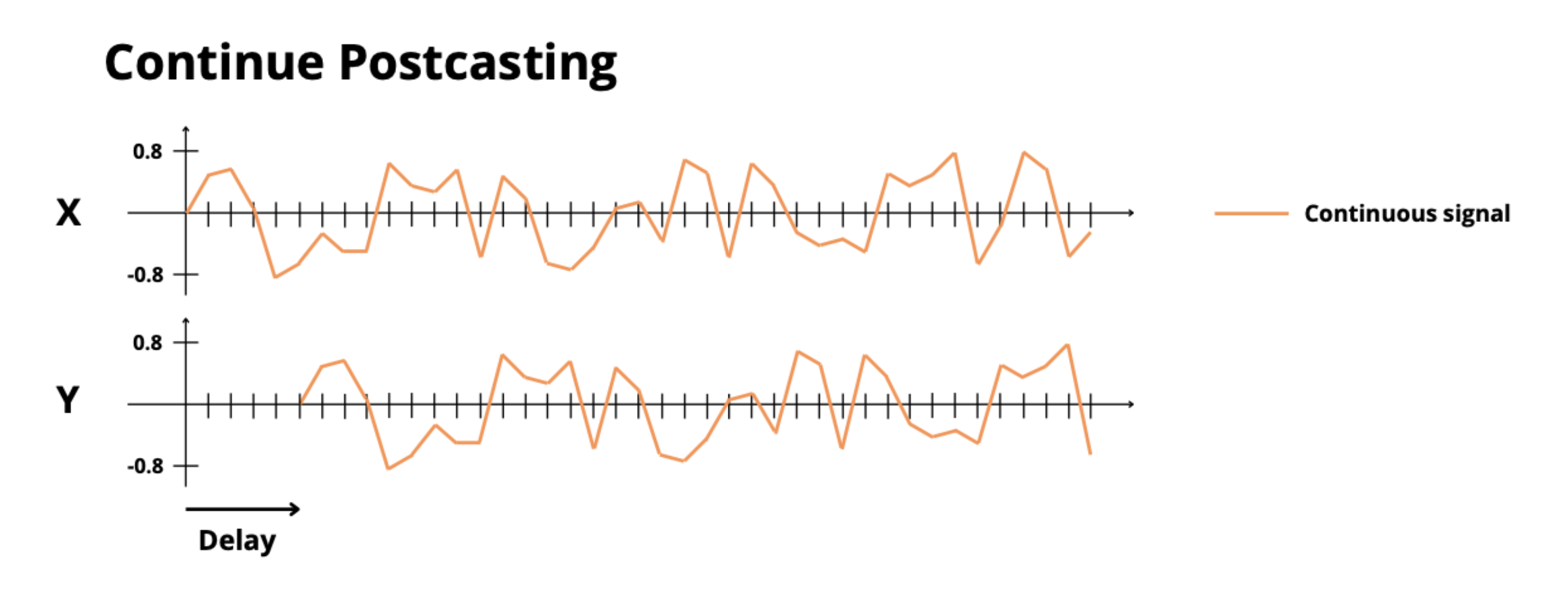}
\caption{Continuous Postcasting}
\label{fig:continuous_postcasting}
\end{figure}

\begin{figure}[h]
\centering
\includegraphics[width=0.9\linewidth]{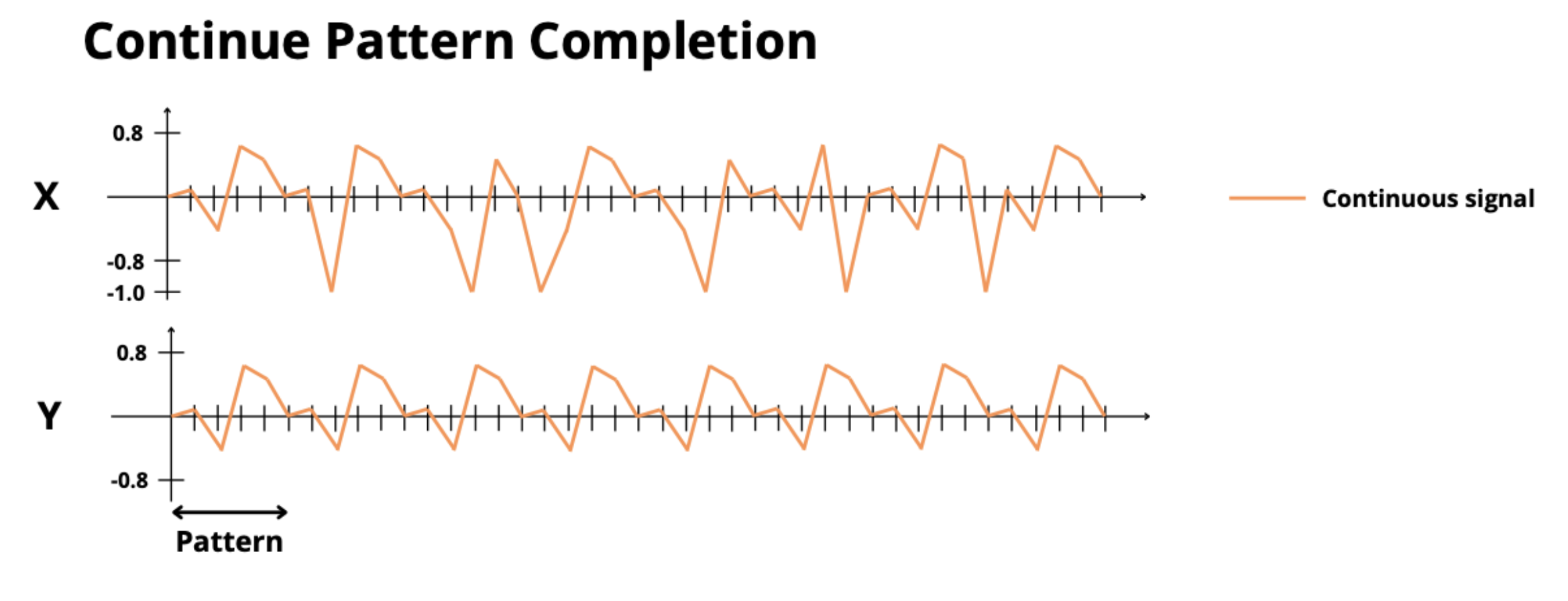}
\caption{Continuous Pattern Completion}
\label{fig:continuous_pattern_completion}
\end{figure}

\begin{figure}[h]
\centering
\includegraphics[width=0.9\linewidth]{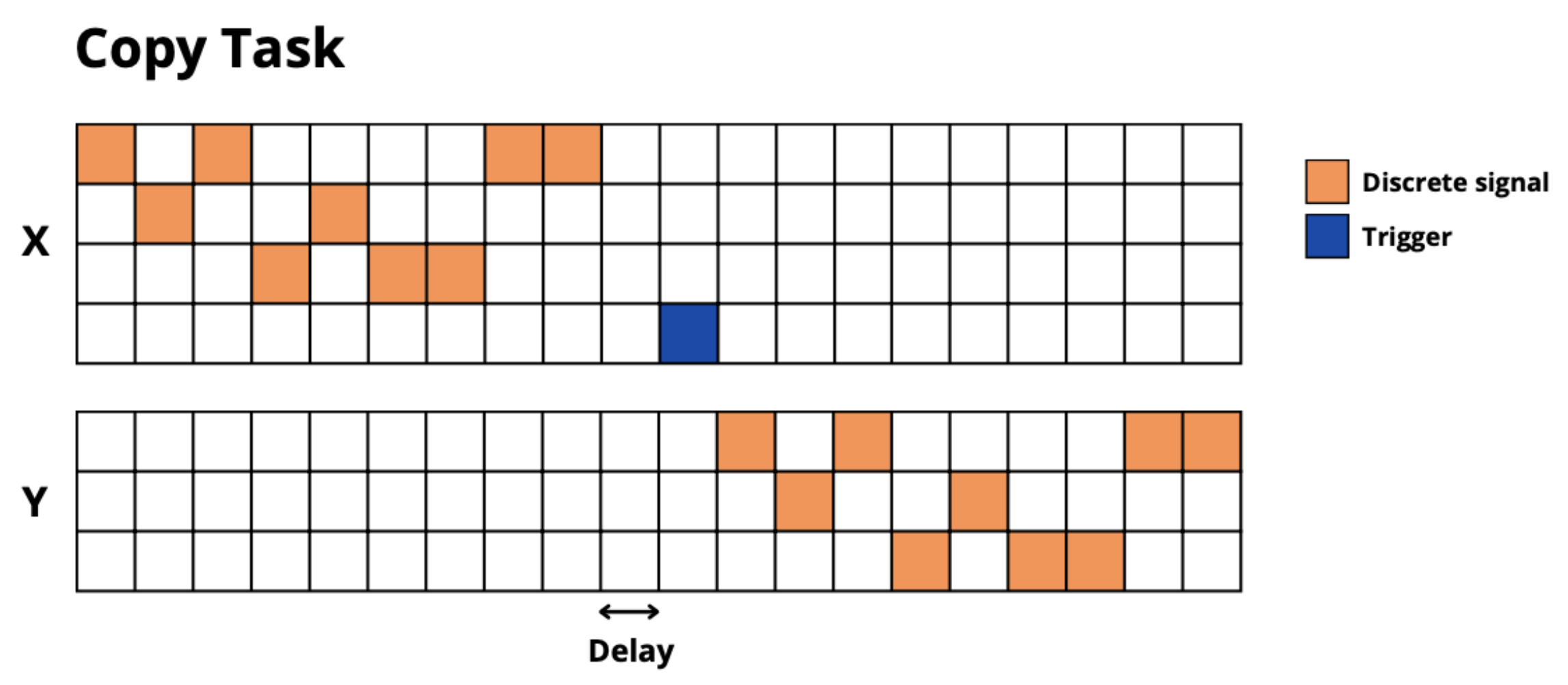}
\caption{Simple copy}
\label{fig:simple_copy}
\end{figure}

\begin{figure}[h]
\centering
\includegraphics[width=0.9\linewidth]{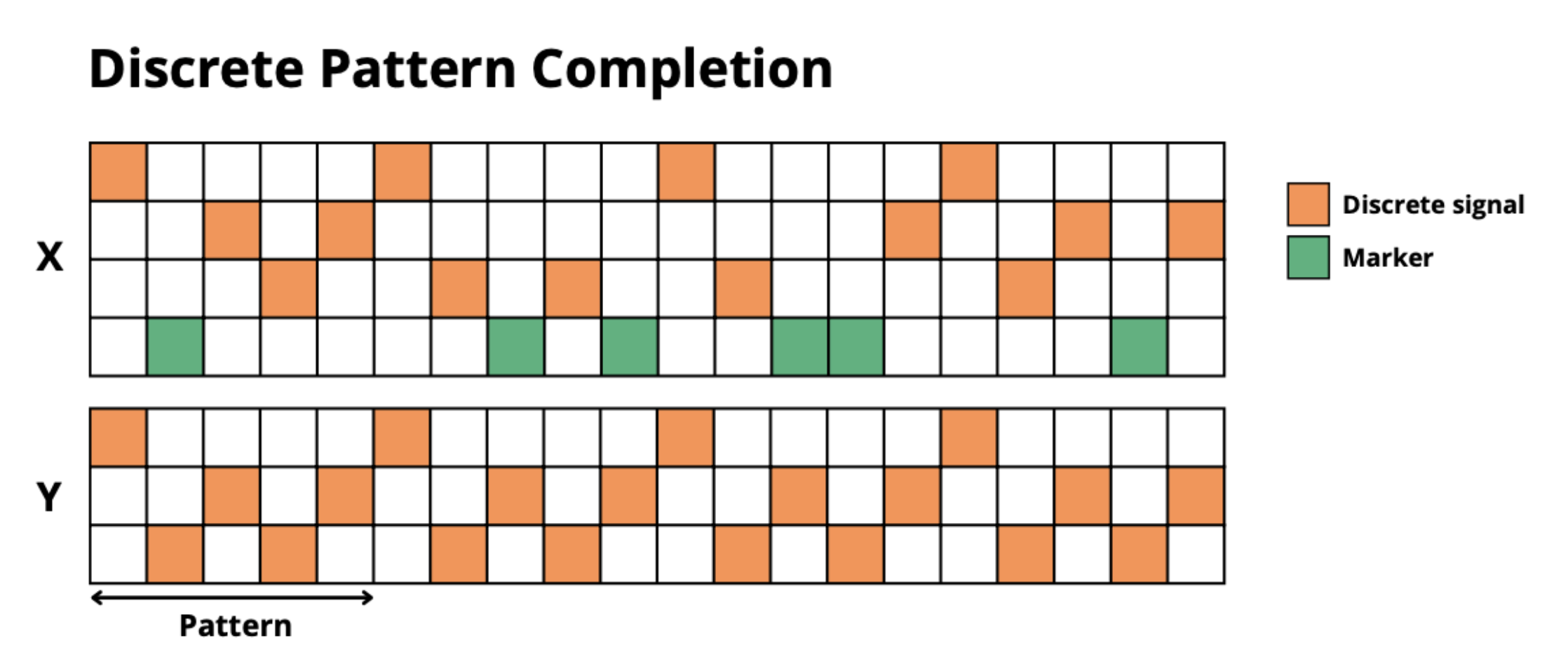}
\caption{Discrete pattern completion}
\label{fig:dis_pat_comp}
\end{figure}

\begin{figure}[h]
\centering
\includegraphics[width=0.9\linewidth]{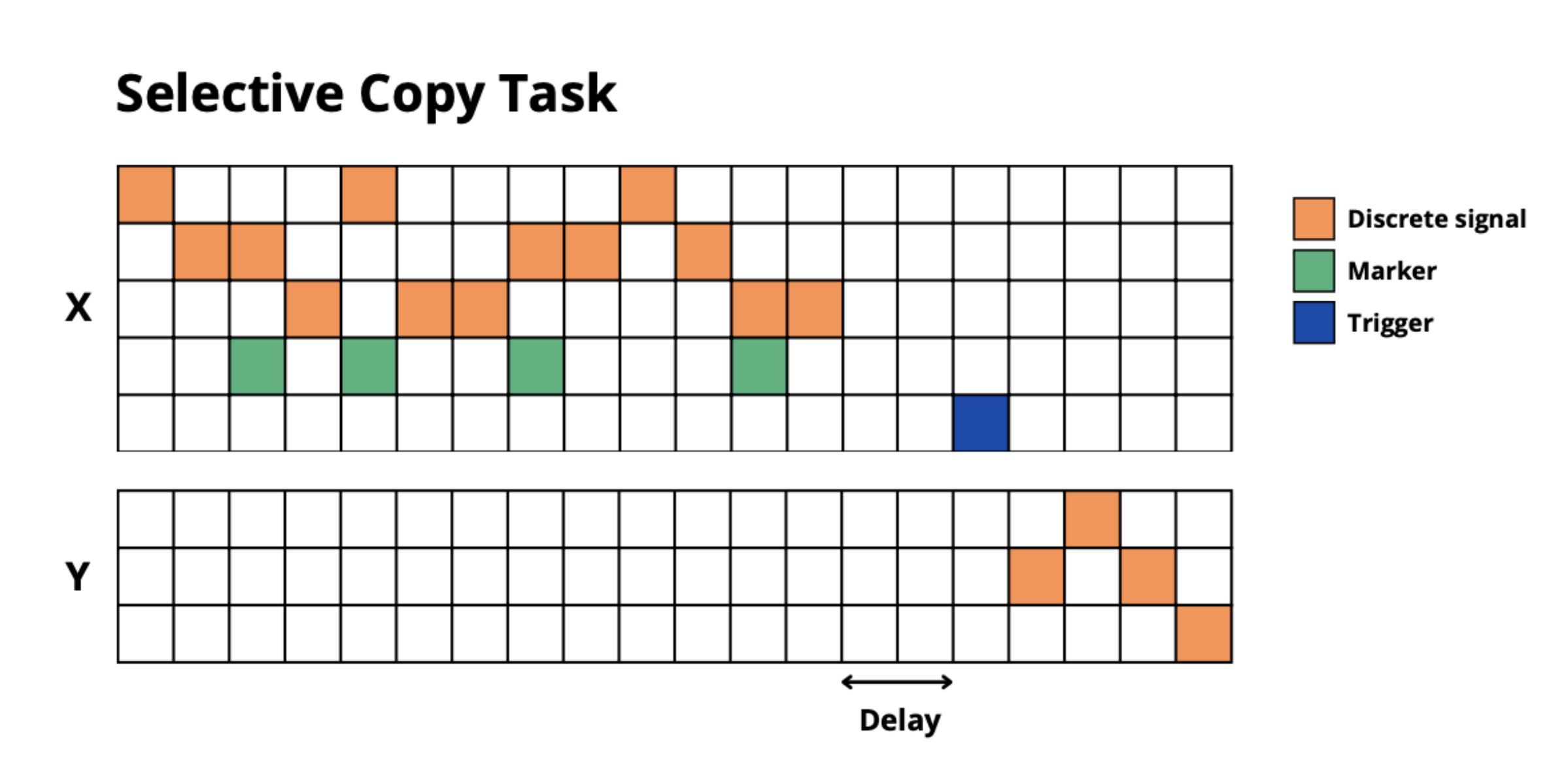}
\caption{Selective copy}
\label{fig:selective_copy}
\end{figure}

\begin{figure}[h]
\centering
\includegraphics[width=0.9\linewidth]{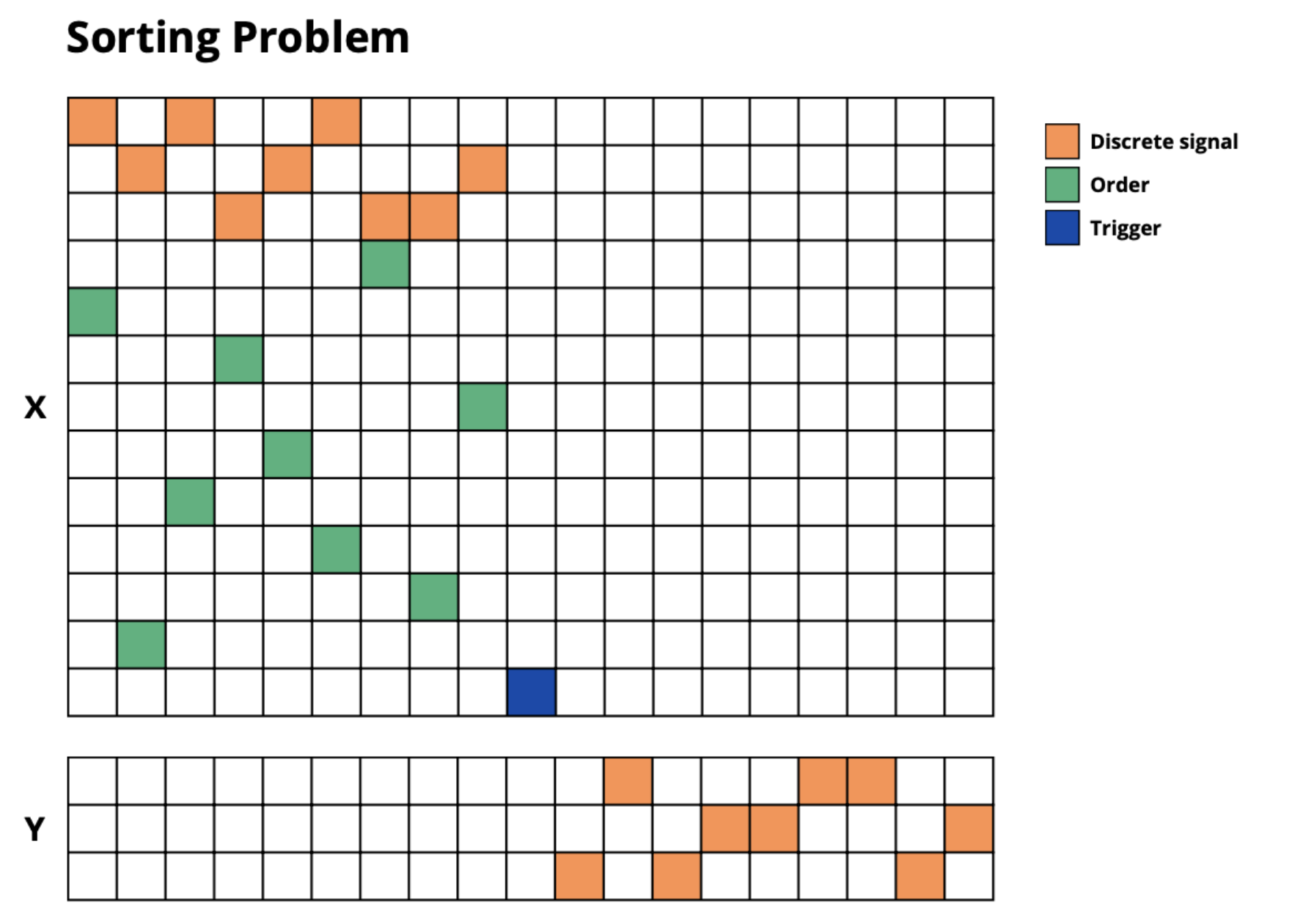}
\caption{Sorting problem}
\label{fig:sorting_problem}
\end{figure}

\begin{figure}[h]
\centering
\includegraphics[width=0.9\linewidth]{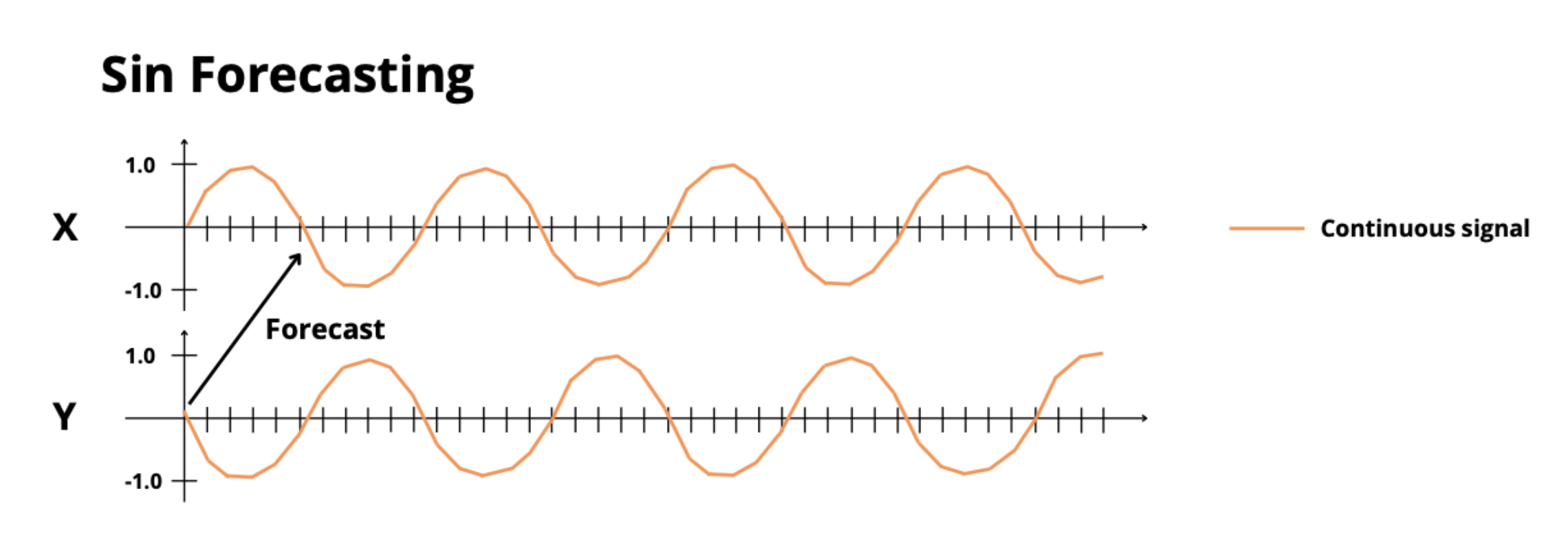}
\caption{Sinus forecasting}
\label{fig:sinus_forecasting}
\end{figure}

\clearpage
\section{EARLY and Random Search Generalisation Heat Maps Comparison}
\begin{figure}[h]
\centering
\includegraphics[width=0.9\linewidth]{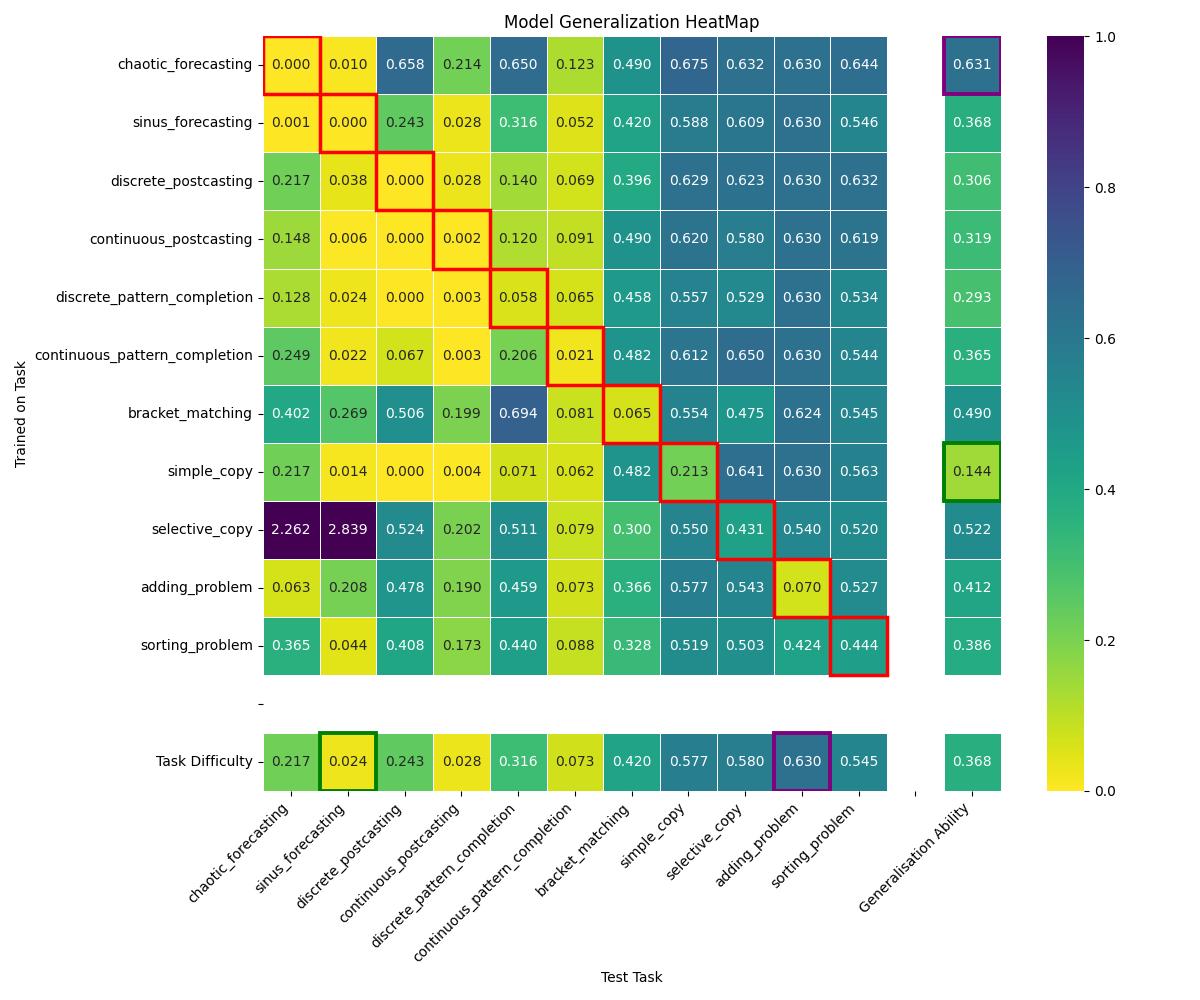}
\caption{HeatMap EARLY}
\label{fig:HM_EARLY}
\end{figure}

\begin{figure}[h]
\centering
\includegraphics[width=0.9\linewidth]{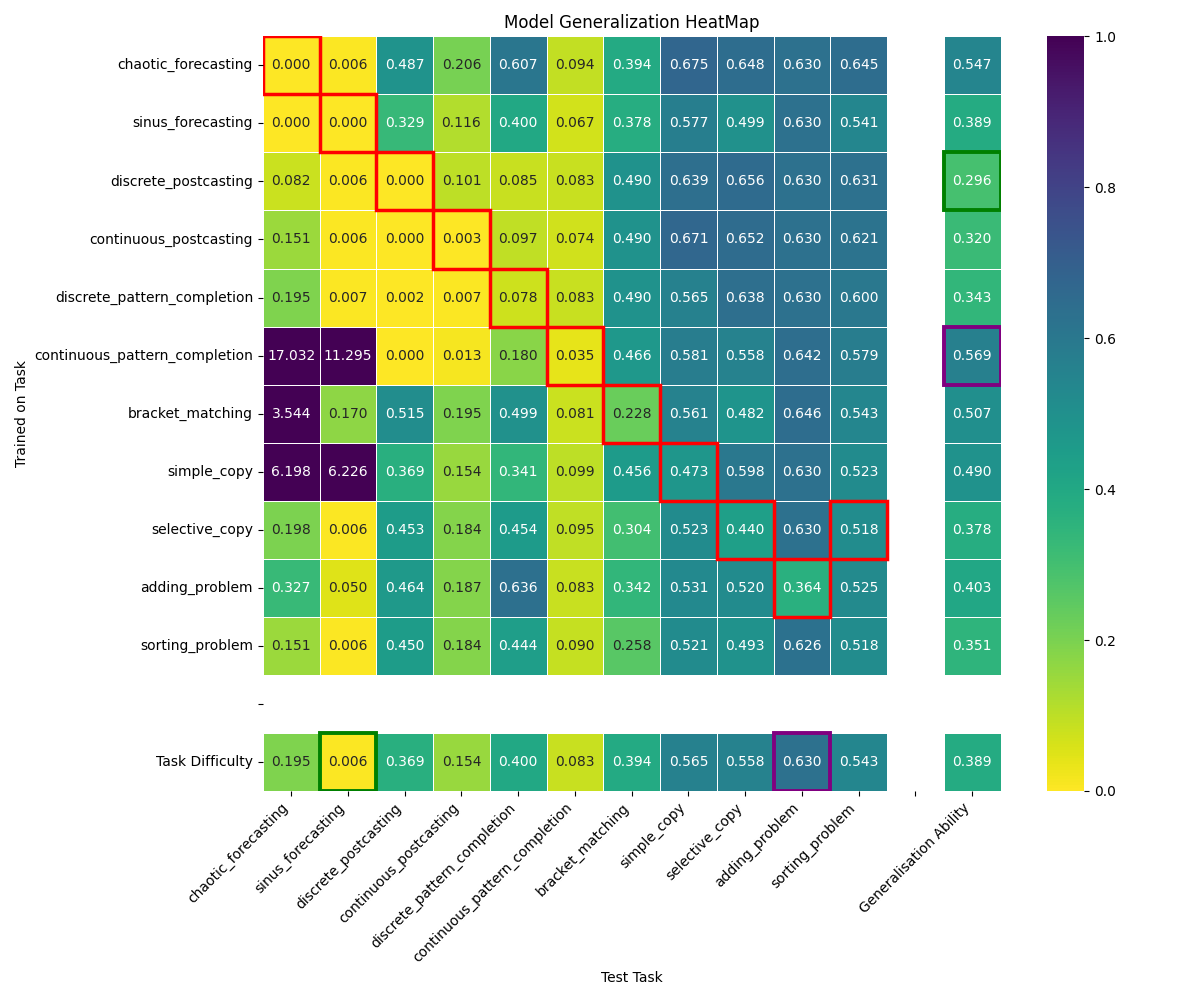}
\caption{HeatMap Random Search}
\label{fig:HM_RS}
\end{figure}

\end{document}